\documentclass[journal,twoside,web]{ieeecolor}
\usepackage{generic}
\usepackage{cite}
\usepackage[utf8]{inputenc}
\usepackage[T1]{fontenc}
\usepackage{amsmath,amsfonts,amssymb}
\usepackage{booktabs}
\usepackage{graphicx}
\usepackage{url}
\usepackage{xcolor}
\definecolor{subsectioncolor}{rgb}{0.075,0.541,0.855}
\usepackage{colortbl}
\usepackage{capt-of}
\usepackage{algorithm}
\usepackage{algpseudocode}
\usepackage{framed}
\usepackage{listings}
\usepackage{multirow}
\usepackage{array}
\usepackage{tabularx}
\usepackage{placeins}
\usepackage{needspace}
\usepackage{balance}
\usepackage[normalem]{ulem}
\makeatletter
\let\NAT@parse\undefined
\makeatother
\usepackage{hyperref}
\usepackage[all]{hypcap}
\hypersetup{
  colorlinks=true,
  linkcolor=subsectioncolor,
  citecolor=blue,
  urlcolor=subsectioncolor,
  filecolor=subsectioncolor,
  hypertexnames=false,
  pdftitle={PanDent: Toward Comprehensive Tooth-Level Structure-Language Consistency in Dental Radiology},
  pdfauthor={Xiaohan Li, Xinyu Liu, Chang Liu, Kuofeng Hung, Sum Wing Au Yeung, Jun Liu, Yixuan Yuan, and Hui Chen}
}

\newcommand{\TMIfigref}[1]{\hyperref[#1]{Fig.~\ref*{#1}}}
\newcommand{\TMIfigrefs}[2]{\hyperref[#1]{Figs.~\ref*{#1}} and~\hyperref[#2]{\ref*{#2}}}
\newcommand{\TMItabref}[1]{\hyperref[#1]{Table~\ref*{#1}}}

\def\BibTeX{{\rm B\kern-.05em{\sc i\kern-.025em b}\kern-.08em
    T\kern-.1667em\lower.7ex\hbox{E}\kern-.125emX}}

\providecommand{\refname}{References}

\graphicspath{{figures/}{./}}


\newcommand{\PanDentAuthorNames}{%
  Xiaohan Li, Xinyu Liu, Chang Liu, Kuofeng Hung,\\
  Sum Wing Au Yeung, Jun Liu, Yixuan Yuan, and Hui Chen}

\algrenewcommand\algorithmicrequire{\textbf{Input:}}
\algrenewcommand\algorithmicensure{\textbf{Output:}}

\algrenewcommand\algorithmiccomment[1]{%
  \hfill\textcolor{green!50!black}{$\triangleleft$~\footnotesize #1}%
}

\newcommand{\posdelta}[2]{\ensuremath{#1_{\scriptscriptstyle\textcolor{green!60!black}{+#2}}}}
\newcommand{\negdelta}[2]{\ensuremath{#1_{\scriptscriptstyle\textcolor{red!70!black}{-#2}}}}
\newcommand{\best}[1]{\cellcolor{red!12}#1}
\newcommand{\second}[1]{\cellcolor{blue!10}#1}

\newcommand{\nasize}{\textcolor{gray!70}{N/A}}
\definecolor{DentalHit}{HTML}{137A52}
\definecolor{DentalMiss}{HTML}{B4433E}
\definecolor{DentalAttr}{HTML}{315D9A}
\definecolor{DentalModelBg}{HTML}{EEF7F1}
\definecolor{CaseMissingBg}{HTML}{F1E7FA}
\definecolor{CasePathBg}{HTML}{FBE4E2}
\definecolor{CaseTreatmentBg}{HTML}{E7F0FF}
\definecolor{CaseBoneBg}{HTML}{FFF0D6}
\definecolor{CaseAnatomyBg}{HTML}{EAF6EA}
\definecolor{CaseRawBg}{HTML}{F2F2F2}
\definecolor{CaseNormBg}{HTML}{EEF4FF}

\newcommand{\reportsep}{\par\noindent\textcolor{gray!35}{\rule{\linewidth}{0.25pt}}\par}
\newcommand{\modeldentbox}[1]{\begingroup\colorbox{DentalModelBg}{\parbox{\dimexpr\linewidth-2\fboxsep}{\raggedright #1}}\endgroup}
\newcommand{\casechip}[2]{\begingroup\colorbox{#1}{\strut #2}\endgroup}

\newcommand{\casemissing}[1]{\casechip{CaseMissingBg}{#1}}

\newcommand{\casetreat}[1]{\casechip{CaseTreatmentBg}{#1}}
\newcommand{\casebone}[1]{\casechip{CaseBoneBg}{#1}}
\newcommand{\caseanat}[1]{\casechip{CaseAnatomyBg}{#1}}

\newlength{\datasetrowheight}
\newcommand{\casegood}[1]{\textcolor{DentalHit}{\uline{\textbf{#1}}}}
\newcommand{\casebad}[1]{\textcolor{DentalMiss}{\uline{\textbf{#1}}}}
\newcommand{\caseattrbox}[1]{\begingroup\colorbox{gray!8}{\parbox{\dimexpr\linewidth-2\fboxsep}{\emergencystretch=1em #1}}\endgroup}
\newcommand{\caserawbox}[1]{\begingroup\colorbox{CaseRawBg}{\parbox{\dimexpr\linewidth-2\fboxsep}{\fontsize{5.8}{5.9}\selectfont\emergencystretch=1em\raggedright #1}}\endgroup}
\newcommand{\casenormbox}[1]{\begingroup\colorbox{CaseNormBg}{\parbox{\dimexpr\linewidth-2\fboxsep}{\fontsize{5.8}{5.9}\selectfont\emergencystretch=1em\raggedright #1}}\endgroup}
\newcommand{\caserawnorm}[2]{\par\noindent\begingroup\begin{tabularx}{\linewidth}{@{}>{\hsize=0.6\hsize\linewidth=\hsize\raggedright\arraybackslash}X@{{\scriptsize$\Rightarrow$}}>{\hsize=1.4\hsize\linewidth=\hsize\raggedright\arraybackslash}X@{}}\caserawbox{#1} & \casenormbox{#2}\end{tabularx}\endgroup}
\definecolor{promptblue}{RGB}{35,88,220}
\lstdefinestyle{promptstyle}{
  basicstyle=\ttfamily\scriptsize,
  columns=fullflexible,
  keepspaces=true,
  breaklines=true,
  breakatwhitespace=false,
  showstringspaces=false,
  backgroundcolor=\color{gray!10},
  frame=single,
  rulecolor=\color{gray!25},
  framerule=0pt,
  xleftmargin=4pt,
  xrightmargin=4pt,
  escapeinside={(*@}{@*)}
}

\begin{document}
\flushbottom

\bstctlcite{IEEEtranBSTcontrol}
\title{\textsc{PanDent}: Toward Comprehensive Tooth-Level Structure-Language Consistency in Dental Radiology}
\author{\PanDentAuthorNames%
\thanks{This work was supported by The University of Hong Kong Seed Fund for Collaborative Research 2024/25 under Project 2407102725. (Corresponding authors: Hui Chen; Yixuan Yuan.)}%
\thanks{Xiaohan Li, Kuofeng Hung, Sum Wing Au Yeung, and Hui Chen are with the Faculty of Dentistry, The University of Hong Kong, Hong Kong, China (e-mail: li2xh@hku.hk; 
hungkfg@hku.hk; sharonay@connect.hku.hk; amyhchen@hku.hk).}%
\thanks{Xinyu Liu is with the Department of Computing, Imperial College London, London, United Kingdom (e-mail: xinyuliu@link.cuhk.edu.hk).}%
\thanks{Chang Liu is with the University of Science and Technology of China, Hefei, China (e-mail: lc980413@mail.ustc.edu.cn).}%
\thanks{Jun Liu is with the Department of Data and Systems Engineering, The University of Hong Kong, Hong Kong, China (e-mail: djliu@hku.hk).}%
\thanks{Yixuan Yuan is with the Department of Electronic Engineering, The Chinese University of Hong Kong, Hong Kong, China (e-mail: yxyuan@ee.cuhk.edu.hk).}%
\thanks{Xiaohan Li, Xinyu Liu, and Chang Liu contributed equally to this work.}}

\maketitle

\begin{abstract}
Accurate evaluation of multimodal large language models in dental panoramic radiography is limited by the lack of fine-grained, clinically reliable benchmarks that reflect expert interpretation. This work introduces \textsc{PanDent}, a large-scale benchmark built on expert-validated tooth-level annotations. It comprises $9{,}554$ panoramic radiographs with comprehensive structured findings produced by experienced dentists and further validated by an oral and maxillofacial radiologist. Clinician-guided reports are constructed from these findings using clinician-defined reporting logic, establishing explicit correspondence between structured clinical evidence and free-text descriptions. The dataset supports complementary evaluation of report-level textual agreement and tooth-level clinical consistency, with $9{,}049$ public-source cases for supervised training and an independent $505$-case in-house benchmark. Experiments cover proprietary, general-domain open-source, and medical-specific models. Current models generate fluent reports but exhibit substantial errors in fine-grained localization and tooth-level diagnosis. Supervised fine-tuning on \textsc{PanDent} markedly improves report alignment and diagnostic accuracy, bringing predictions closer to expert dental interpretation. These results establish \textsc{PanDent} as a rigorous benchmark and training resource for clinically grounded dental multimodal reasoning. Dataset sample: {\urlstyle{same}\nolinkurl{https://huggingface.co/datasets/Desperado1103/Pandent_Sample}}; code: {\urlstyle{same}\nolinkurl{https://github.com/HKUDentistry/PanDent}}. Full data will be released upon publication under source redistribution terms.
\end{abstract}

\begin{IEEEkeywords}
dental panoramic radiography, multimodal large language models, medical image analysis, tooth-level reasoning, visual grounding, clinical diagnosis
\end{IEEEkeywords}

\section{Introduction}

Dental panoramic radiography (orthopantomogram, OPG) is widely used in routine dental practice, providing a comprehensive view of the dentition and surrounding anatomical structures for radiographic evaluation, diagnosis, and treatment planning~\cite{SinghRaza2022DentalReview,Panetta2022TuftsDental}. In daily clinical practice, OPGs are interpreted by dental professionals and documented in free-text notes. However, because manual report generation is highly labor-intensive, these notes are typically restricted to addressing the patient's primary complaint. Clinicians rarely write exhaustive reports detailing all radiographic findings, a practice that often results in omitted incidental findings and missed secondary diagnoses. Moreover, the manual interpretation of OPGs is highly susceptible to significant inter-observer variability, particularly when complex, co-occurring conditions must be jointly evaluated. In addition, OPGs are two-dimensional projections of complex three-dimensional maxillofacial anatomy and are affected by geometric distortion, anatomical superposition, and scale ambiguity. These factors make fine-grained interpretation challenging even for experienced clinicians.

Recent work has explored data-driven models for extracting clinical findings and generating structured or free-text descriptions from OPG images. However, progress is limited by the lack of large, clinically reliable benchmarks for expert-level evaluation~\cite{lutescu2026can}. Existing public benchmarks for medical multimodal large language models (MLLMs) are predominantly centered on chest X-ray imaging, including report-generation datasets such as MIMIC-CXR~\cite{Johnson2019MIMICCXR}, IU X-Ray~\cite{DemnerFushman2016OpenI}, and PadChest~\cite{Bustos2020PadChest}, as well as medical visual question answering benchmarks such as ImageCLEF-VQA-Med~\cite{BenAbacha2019VQAMed}, SLAKE~\cite{Liu2021SLAKE}, and PathVQA~\cite{He2020PathVQA}. While these datasets have advanced medical vision-language modeling, they primarily focus on global image-level understanding and do not reflect the structure of dental panoramic interpretation. 

In OPG images, clinically meaningful reasoning depends on accurate tooth-level localization and interpretation of surrounding anatomical structures. However, existing dental datasets predominantly focus on isolated recognition tasks, such as tooth segmentation, numbering, or narrow-scope abnormality classification. Although some datasets provide tooth-level annotations, their scope is typically restricted to a small set of task-specific findings and does not jointly capture the broad range of tooth-specific and regional information required for comprehensive clinical interpretation. In other datasets, abnormalities may be represented only at the image level; for example, dental caries can be annotated by overall presence rather than by the specific teeth affected. Such supervision is insufficient for evaluating whether models can accurately localize diverse clinical findings and associate them with the corresponding anatomical structures. Consequently, existing datasets offer inadequate support for assessing model outputs against expert dental interpretation, making it difficult to distinguish genuine clinical understanding from linguistically plausible reporting. In clinical reporting, structured findings and language are tightly coupled: clinicians first identify findings at specific teeth and then express them through structured yet flexible diagnostic descriptions. Benchmarks must therefore model structure--language consistency to support reliable evaluation and develop models that approach expert dental interpretation.

Building on this motivation, \textsc{PanDent} is introduced as a large-scale, clinically grounded benchmark for evaluating fine-grained tooth-level understanding and structure-language consistency in panoramic dental imaging. Starting from $9{,}854$ raw annotated records, $9{,}554$ high-quality radiograph-report pairs are curated, including $9{,}049$ public-source cases for model training and an independent $505$-case in-house benchmark. Each case is associated with comprehensive structured annotations spanning tooth-specific and regional clinical findings. To the best of our knowledge, it is among the first large-scale datasets that comprehensively cover a broad range of clinically relevant findings observable in OPG images. Four experienced dentists produce these annotations, which are further validated by an expert in oral and maxillofacial radiology. Clinician-defined reporting rules then map the findings to reference descriptions, preserving tooth-level fidelity and making each diagnostic statement traceable to its source annotation. Reference content is anchored in clinician annotations, rather than independently generated by a model. 

\textsc{PanDent} is designed to serve both as an evaluation benchmark and as a training resource. As a benchmark, it enables explicit assessment of whether model outputs are consistent with expert-validated tooth-level findings, going beyond conventional report-level metrics that mainly measure fluency or textual similarity. As a training resource, its scale and structured supervision support supervised fine-tuning for improving dental-domain multimodal reasoning. Accordingly, experiments are organized around three complementary aspects: report-level textual agreement, attribute-level visual diagnostic accuracy, and performance improvement through supervised fine-tuning. Results suggest that current MLLMs can generate fluent reports while still making substantial errors in fine-grained clinical findings, whereas fine-tuning on \textsc{PanDent} significantly improves dental-domain performance and brings model outputs closer to expert-validated annotations. 

In summary, the contributions are three-fold:
\begin{itemize}
\item \textsc{PanDent} is presented as a large-scale dental panoramic benchmark with $9{,}554$ radiograph-report pairs and comprehensive, expert-validated tooth-level annotations, covering a broad range of clinically relevant findings.
\item The dataset links clinician-derived tooth-level and regional annotations with corresponding reports, supporting complementary evaluation of textual agreement and clinical consistency through findings extracted from the same generated reports.
\item A comprehensive evaluation across mainstream closed-source, open-source, and medical-domain MLLMs reveals that fluent report generation does not imply clinically consistent tooth-level reasoning, and demonstrates that fine-tuning on \textsc{PanDent} significantly improves diagnostic performance, bringing model outputs closer to expert-level dental interpretation.
\end{itemize}

\begin{figure*}[t]
    \centering
    \includegraphics[width=\textwidth]{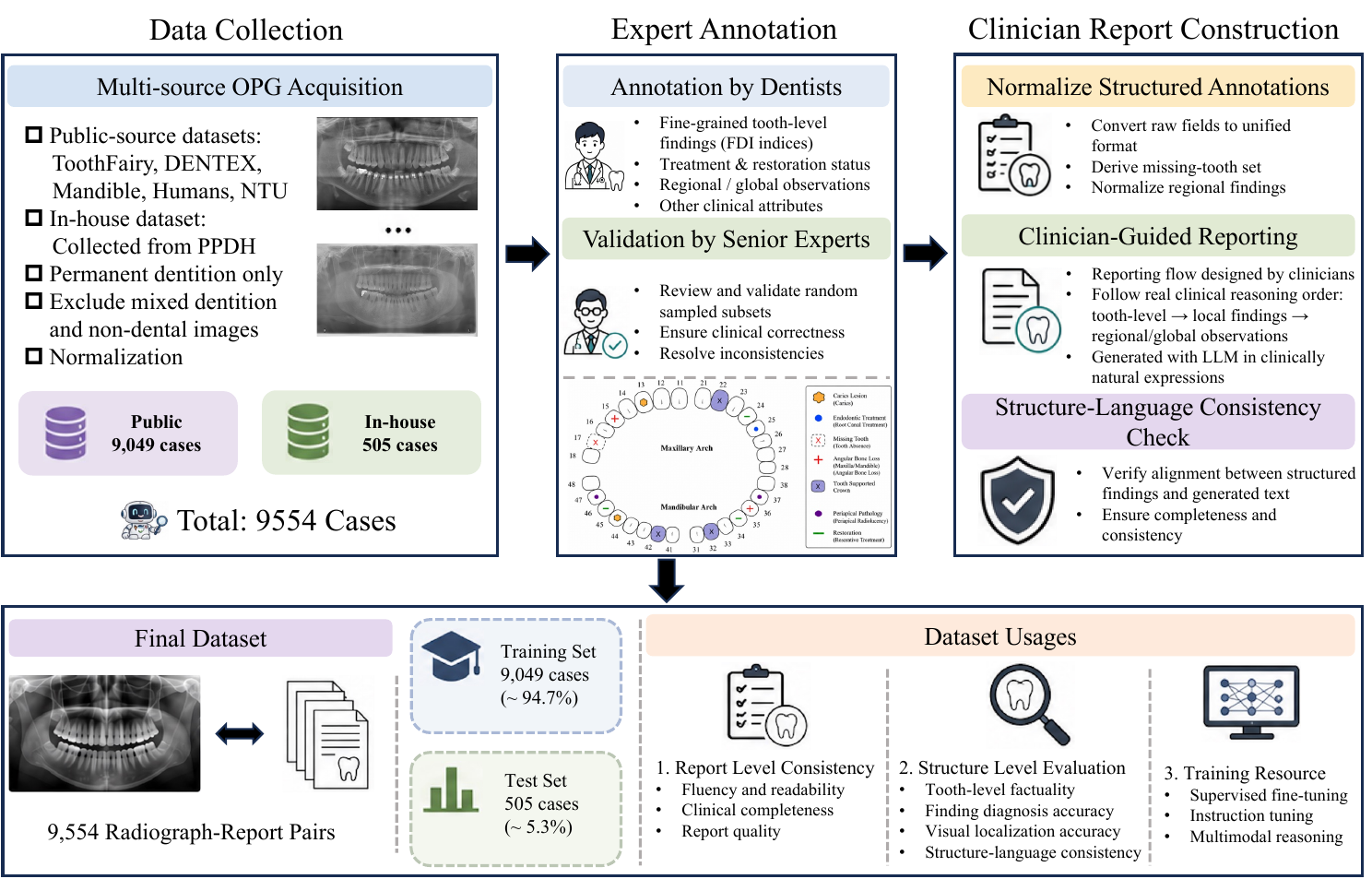}
    \caption{\textsc{PanDent} pipeline: multi-source OPG collection, expert-validated tooth-level findings, clinician-guided reports, and dataset release.}
    \label{fig:dataset_construction}
\end{figure*}

\section{Related Works}
\label{sec:related-works}
\paragraph{Medical Radiology Datasets}
Medical radiology datasets provide complementary supervision through image--report pairs, image-grounded question--answer pairs, and structured clinical labels. IU X-Ray~\cite{DemnerFushman2016OpenI} and MIMIC-CXR~\cite{Johnson2019MIMICCXR} pair chest radiographs with diagnostic reports, with MIMIC-CXR offering a large de-identified corpus. PadChest~\cite{Bustos2020PadChest} adds report-derived multi-label annotations at the image level. ImageCLEF-VQA-Med~\cite{BenAbacha2019VQAMed}, SLAKE~\cite{Liu2021SLAKE}, and PathVQA~\cite{He2020PathVQA} support targeted medical image reasoning through question--answer pairs. RadGraph~\cite{Jain2021RadGraph} represents report-derived clinical entities and relations for factual evaluation. These forms respectively support automatic reporting~\cite{gao2026s2dalign}, targeted image-grounded reasoning, and factual analysis of generated text. They advance medical image understanding and language generation, but their annotation schemes are not tailored to the tooth-index-aware findings and terminology required for panoramic dental interpretation.

\paragraph{Dental Radiology Datasets}
Public dental radiology datasets have primarily supported visual recognition tasks. The Tufts Dental Database~\cite{Panetta2022TuftsDental} combines panoramic radiographs with anatomical and abnormality annotations and auxiliary eye-tracking information for studying expert interpretation. OdontoAI~\cite{Silva2023OdontoAI} provides tooth segmentation and numbering annotations, together with an online platform for benchmarking these tasks. DENTEX~\cite{Hamamci2023DENTEX} introduces hierarchical annotations for dental enumeration, abnormal-tooth detection, and diagnosis. Related research also addresses intraoral radiographs and cone-beam computed tomography for caries screening, periodontal assessment, landmark localization, and lesion segmentation~\cite{SinghRaza2022DentalReview}. These resources support the recognition and localization of anatomical structures or selected abnormalities, but are not primarily designed to associate a comprehensive set of clinical findings with complete radiology reports.

More recently, MMOral~\cite{Hao2025MMOral} has extended panoramic image analysis to multimodal instruction learning, covering attribute extraction, report generation, visual question answering, and image-grounded dialogue. Clinical findings are extracted using specialist visual models and associated with tooth indices through spatial relationships, after which reports are generated and refined using LLMs. Although report quality is assessed by dentists and the accompanying benchmark undergoes manual validation, the large-scale annotations are primarily model-derived. The published visual extraction taxonomy also does not separately encode several clinical distinctions, including restoration material, post-and-core structures, furcation involvement, and temporomandibular joint abnormalities. By comparison, \textsc{PanDent} is built on clinician-annotated tooth-specific and regional findings, from which reference reports are constructed using clinician-defined reporting rules.

\Needspace{5\baselineskip}
\section{Data Curation}
\label{sec:data_curation}
\TMIfigref{fig:dataset_construction} summarizes the curation pipeline. Removing $300$ records with invalid or unresolved image paths yields $9{,}554$ valid cases: $9{,}049$ public-source and $505$ in-house OPGs. Public sources include ToothFairy (MICCAI 2023), DENTEX (MICCAI 2023), Panoramic Dental X-rays with Segmented Mandibles, Teeth Segmentation on Dental X-ray Images (Humans in the Loop), and NTU Dentistry-Vibir. These sources introduce variation in institutions, imaging equipment, acquisition protocols, and annotation characteristics.

Four dentists, each with over ten years of clinical experience, contribute annotations covering more than $20$ clinically relevant findings. Annotation review follows separate procedures for the public-source and in-house subsets. For the public-source data, an assistant professor in oral and maxillofacial radiology reviews a random sample of $1{,}000$ annotated OPGs for quality assurance. For the in-house benchmark, all $505$ OPGs are independently annotated at the tooth level by two senior clinicians. The same radiologist reviews every in-house case and adjudicates disagreements between the two clinicians to establish the final reference annotations. A separate $100$ case study of public-source OPGs assesses cross-annotator reproducibility. All $9{,}049$ public-source cases are used for training, whereas the in-house benchmark is reserved exclusively for evaluation.

\textbf{Ethics and privacy.}
In-house radiographs are de-identified before processing. Ethical approval was obtained from the Institutional Review Board of The University of Hong Kong/Hospital Authority Hong Kong West Cluster (UW 24--520) before the study commenced. All procedures comply with institutional privacy and ethical-use requirements.

\begin{figure*}[t]
    \centering
    \includegraphics[width=\textwidth]{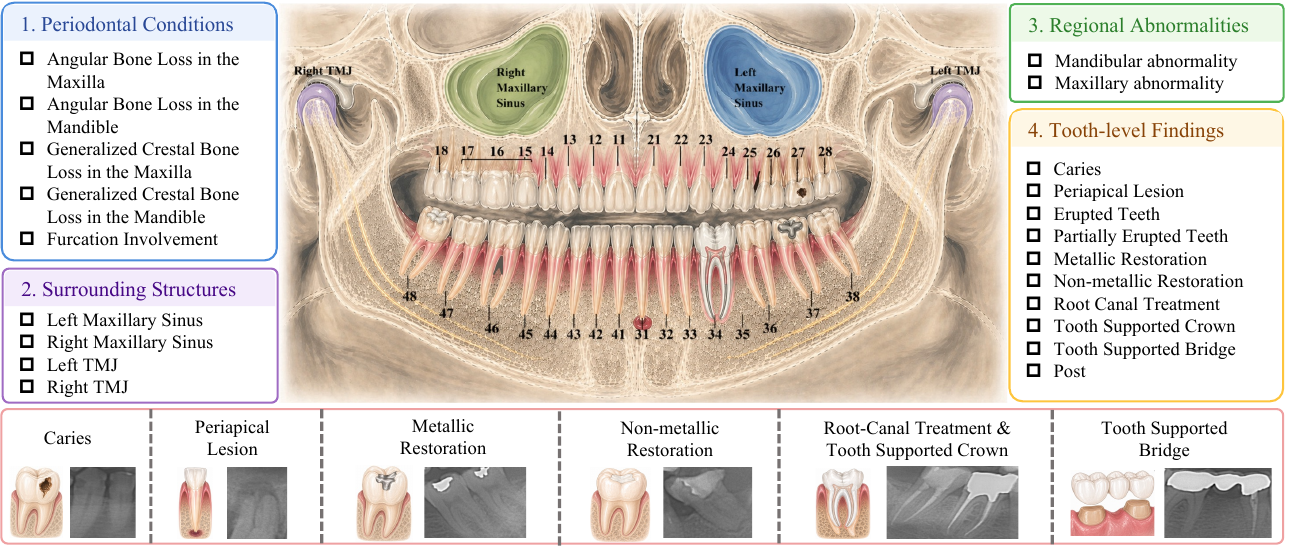}
    \caption{Four-aspect annotation schema: periodontal conditions, surrounding structures, regional abnormalities, and tooth-level findings. Tooth findings use FDI indices; regional findings use anatomical labels. Tooth-level examples are shown below.}
    \label{fig:dataset_illustration}
\end{figure*}

\subsection{Attribute Labeling and Normalization}
\label{sec:attribute-normalization}
As shown in \TMIfigref{fig:dataset_illustration}, raw fields are consolidated into $21$ reportable finding dimensions covering regional abnormalities, structural and periodontal conditions, tooth-level pathologies and treatments, and surrounding anatomical structures. The \texttt{ans\_comment} field captures other observations, including orthodontic brackets and mandibular fixation plates or screws.

\textit{Tooth-index-aware findings} are encoded as sets of standardized FDI indices, allowing multiple affected teeth and co-existing findings at the same tooth. They include erupted and partially erupted/impacted teeth, caries, periapical lesions, furcation involvement, metallic and non-metallic restorations, root-canal treatment, tooth-supported crowns and bridges, post-and-core restorations, and localized angular bone loss in either jaw. This representation retains the anatomical correspondence required for report construction and exact tooth-level evaluation.

\textit{Binary regional findings} encode abnormality presence within predefined regions, including maxillary or mandibular generalized crestal bone loss and structural abnormalities, and left/right maxillary sinus or temporomandibular joint abnormalities. Optional descriptive fields preserve additional anatomical or pathological details.

This dual representation preserves the clinically required granularity at both tooth and regional levels, enabling explicit mapping to textual descriptions and attribute-level evaluation of structure--language consistency.

\paragraph{Derivation of Missing Teeth}
Let $\Omega=\{11,\ldots,18,21,\ldots,28,31,\ldots,38,41,\ldots,48\}$ be the $32$ permanent FDI teeth, $\mathcal{E}$ the fully erupted teeth, and $\mathcal{P}$ the partially erupted or impacted teeth. Their union represents all teeth identified as present under the annotation protocol. Missing teeth are therefore derived as
\[
\mathcal{M}=\Omega\setminus(\mathcal{E}\cup\mathcal{P}).
\]
Reports describe $\mathcal{M}$ rather than enumerate all present teeth, restoring the \texttt{\#} prefix when verbalizing FDI indices. This set also constrains other tooth-level findings to prevent assignment to absent teeth. Regional fields are normalized separately: sinus and temporomandibular joint abnormalities retain their laterality, whereas generalized crestal bone loss is recorded at the jaw level. Optional descriptions preserve details that cannot be fully captured by a binary flag.

\subsection{Cross-Annotator Validation}
\label{sec:cross-annotator-validation}
Three of the four primary dentists participate in the reproducibility study. A third annotator independently reassesses $100$ randomly sampled public-source OPGs originally labeled by either of two annotators, recording image-level presence/absence for nine findings prioritized by the radiology expert. This binary review assesses attribute-level reproducibility, not complete tooth-index agreement. Original tooth-set labels are mapped to presence when non-empty; the composite bridge-or-crown finding is positive if either constituent is positive. Raw agreement, Cohen's $\kappa$, and positive-label Jaccard similarity are calculated per attribute. Case-level bootstrap resampling preserves within-image dependence when estimating confidence intervals for pooled agreement and $\kappa$; neither annotation set serves as a reference standard.


\begin{algorithm}[t]
\caption{Clinician-Guided Reference Report Generation}
\label{alg:template-generation}
\begin{algorithmic}[1]
\Require Verified tooth sets $\{S_k\}$ and regional fields $\{(b_k,d_k)\}$
\Require Clinician-defined reporting rules $\mathcal{R}$, comment $c$
\Require LLM report generator $G_{\mathrm{LLM}}$
\Ensure Reference report $Y_{\mathrm{ref}}$
\State $\mathcal{A} \gets \bigl(\{S_k\}, \{(b_k,d_k)\}, c\bigr)$
\State Construct prompt $p$ from $\mathcal{A}$ and $\mathcal{R}$
\State Instruct the LLM to preserve finding polarity, FDI indices, and regional details
\State Instruct the LLM to follow the reporting order without adding unsupported findings
\State $Y_{\mathrm{ref}} \gets G_{\mathrm{LLM}}(p)$
\State \Return $Y_{\mathrm{ref}}$
\end{algorithmic}
\end{algorithm}

\begin{figure*}[t!]
    \centering
    \includegraphics[width=\textwidth]{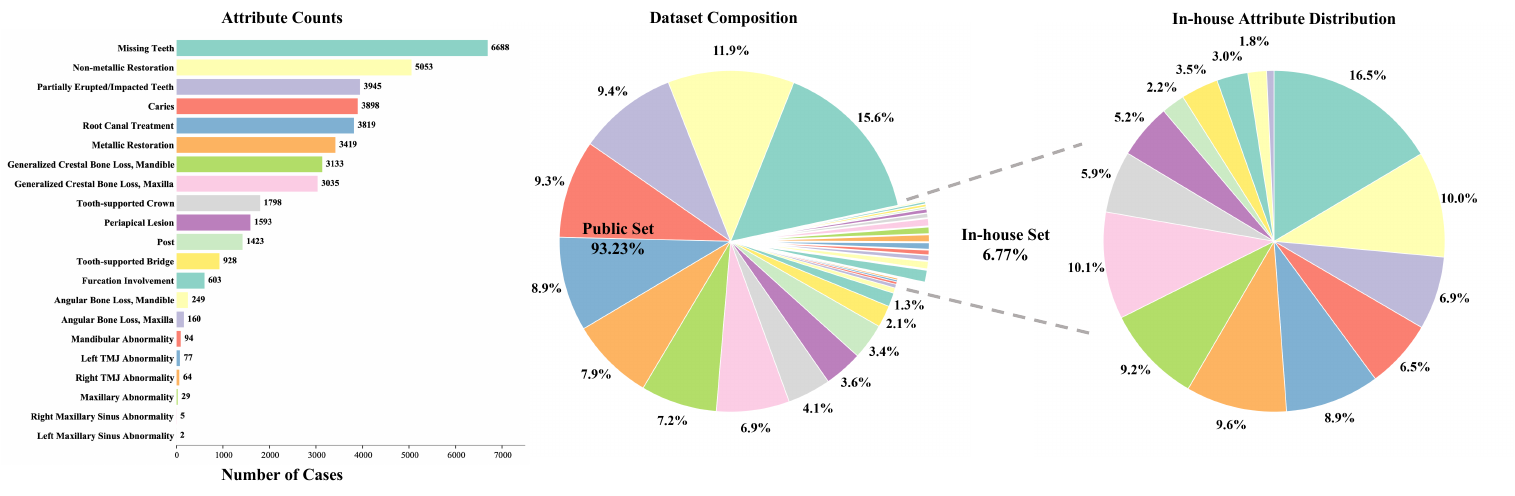}
    
    \caption{Dataset composition and attribute distributions. Left: positive case counts by attribute. Middle: source and attribute composition of positive case--attribute pairs. Right: attribute shares within the in-house subset. Pie charts label proportions above $1\%$.}

    \label{fig:attribute_distribution}
    
\end{figure*}

\begin{table*}[t!]
\centering
\renewcommand{\arraystretch}{1.05}

\caption{Image-level cross-annotator agreement on $100$ public-source OPGs (nine findings; $900$ pooled decisions). Jaccard uses positive labels. $\Delta$ positive rate: independent minus original (percentage points). Independent review combines metallic and non-metallic posts under Post. The bridge-or-crown finding is evaluated jointly here, whereas crown is evaluated separately in \TMItabref{tab:attr_detailed_results}.}
\label{tab:cross-annotator-validation}

\resizebox{0.96\textwidth}{!}{%
\begin{tabular}{lcccccc}
\toprule
\textbf{Attribute} &
\textbf{Original Positive} &
\textbf{Independent Positive} &
\textbf{Agreement (\%)} &
\textbf{Cohen's $\kappa$} &
\textbf{Jaccard} &
\textbf{$\Delta$ Positive Rate (pp)} \\
\midrule

Caries
& $51$ & $64$ & $73.0$ & $0.457$ & $0.620$ & $+13$ \\

Mandibular generalized crestal bone loss
& $42$ & $18$ & $64.0$ & $0.198$ & $0.250$ & $-24$ \\

Maxillary generalized crestal bone loss
& $39$ & $16$ & $69.0$ & $0.271$ & $0.279$ & $-23$ \\

Tooth-supported bridge or crown
& $26$ & $21$ & $93.0$ & $0.806$ & $0.741$ & $-5$ \\

Root-canal treatment
& $47$ & $51$ & $96.0$ & $0.920$ & $0.922$ & $+4$ \\

Periapical lesion
& $19$ & $44$ & $73.0$ & $0.417$ & $0.400$ & $+25$ \\

Post
& $18$ & $26$ & $90.0$ & $0.711$ & $0.630$ & $+8$ \\

Metallic restoration
& $33$ & $50$ & $79.0$ & $0.580$ & $0.596$ & $+17$ \\

Non-metallic restoration
& $53$ & $77$ & $74.0$ & $0.463$ & $0.667$ & $+24$ \\

\midrule
\textbf{Overall}
& $328/900$
& $367/900$
& $79.0$
& $0.558$
& $0.572$
& $+4.3$ \\

\bottomrule
\end{tabular}%
}

\end{table*}

\subsection{Clinician-Guided Report Construction}
\label{sec:report-construction}
Four experienced dentists provide positive sentence templates and corresponding negative templates for normal or unremarkable findings across the $21$ reportable dimensions, yielding a bank of $85$ templates. Clinical content is fixed by finding polarity, FDI indices, and available regional details; wording may vary. A fixed random seed controls template selection, and the released reports are frozen for reproducibility.

Reference reports are generated by an LLM from clinician-annotated findings under clinician-defined reporting rules. These rules specify the reporting order and the expression of finding polarity, FDI tooth indices, and regional information. The LLM is instructed to preserve the supplied clinical content while expressing it as a coherent report, without introducing additional findings or altering anatomical assignments. Thus, the clinical reference content originates from clinician annotations, whereas the LLM provides its textual realization. Alg.~\ref{alg:template-generation} summarizes this rule-guided generation process.


\Needspace{4\baselineskip}
\makeatletter
\let\PanDentSubsectionDisplay\thesubsectiondis
\def\thesubsectiondis{\mbox{\textcolor{subsectioncolor}{\PanDentSubsectionDisplay}}}
\makeatother
\subsection{\texorpdfstring{\mbox{\textcolor{subsectioncolor}{Consistency Verification}}}{Consistency Verification}}
\label{sec:consistency-verification}
\let\thesubsectiondis\PanDentSubsectionDisplay
\normalcolor
Verification is applied before and after report construction. \textit{Rule-based checks} first remove tooth-level assignments incompatible with $\mathcal{M}$ and identify fully/partially erupted conflicts, inconsistent flag--detail pairs, invalid FDI lists, and implausible bridge patterns. An \textit{LLM-assisted verifier} then compares rendered reports with structured findings to flag omitted positives, incorrect tooth indices, contradictions, and formatting errors. It acts only as a consistency checker and does not define or modify the clinician-authored reference. Finally, an oral and maxillofacial radiology expert independently audits $200$ randomly sampled reports for clinical accuracy, tooth-level fidelity, and correspondence with the source findings. This review also safeguards against clinically meaningful omissions or inconsistencies missed by automated checks.

\Needspace{5\baselineskip}
\section{Dataset Analysis}
\subsection{Data Statistics}
\label{sec:data-statistics}
Across the $9{,}554$ cases, reference reports average $196.0$ words and $1{,}327.1$ characters. \TMIfigref{fig:attribute_distribution} shows a long-tailed finding distribution across tooth-level and regional attributes. \TMItabref{tab:attribute_prevalence} reports case-level prevalence in the public-source training set, the in-house test set, and the full release. Missing teeth, non-metallic restorations, and caries occur in $70.0\%$, $52.9\%$, and $40.8\%$ of all cases, respectively; temporomandibular joint, maxillary structural, and maxillary sinus abnormalities occur in $1.2\%$, $0.3\%$, and $0.1\%$. Each OPG may contribute to multiple attributes, so case-level prevalence and the attribute-composition percentages in the pie charts use different denominators. Rare findings are retained without artificial oversampling to preserve clinical breadth. Nine prioritized attributes---seven tooth-index-aware and two binary regional findings---are selected for report-derived clinical diagnostic evaluation.













\begin{table*}[t]
\centering
\caption{Case-level prevalence (\%). Train (public)/Test (in-house)/All: 9,049/505/9,554 OPGs; left/right sinus/TMJ findings are combined.}
\label{tab:attribute_prevalence}

\setlength{\tabcolsep}{5pt}
\renewcommand{\arraystretch}{1.08}

\resizebox{0.95\textwidth}{!}{%
\begin{tabular}{@{}lrrrlrrr@{}}
\toprule
\textbf{Attribute} &
\textbf{Train (\%)} &
\textbf{Test (\%)} &
\textbf{All (\%)} &
\textbf{Attribute} &
\textbf{Train (\%)} &
\textbf{Test (\%)} &
\textbf{All (\%)} \\
\midrule
Root-Canal Tx & $39.5$ & $47.9$ & $40.0$
& Tooth Supported Crown & $18.1$ & $31.5$ & $18.8$ \\

Mand. GCBL & $31.9$ & $49.5$ & $32.8$
& Missing Teeth & $69.0$ & $88.3$ & $70.0$ \\

Max. GCBL & $30.5$ & $54.5$ & $31.8$
& Partially Erupted/Impacted & $41.5$ & $37.2$ & $41.3$ \\

Periapical Lesion & $16.0$ & $27.9$ & $16.7$
& Tooth Supported Bridge & $9.2$ & $19.0$ & $9.7$ \\

Post & $15.1$ & $11.9$ & $14.9$
& Furcation Involvement & $5.8$ & $15.8$ & $6.3$ \\

Mand. ABL & $2.2$ & $9.5$ & $2.6$
& TMJ Abn. & $1.2$ & $0.0$ & $1.2$ \\

Max. ABL & $1.6$ & $3.8$ & $1.7$
& Mand. Struct. Abn. & $1.0$ & $0.0$ & $1.0$ \\

Non-metallic Rest. & $52.8$ & $53.9$ & $52.9$
& Max. Struct. Abn. & $0.3$ & $0.0$ & $0.3$ \\

Caries & $41.1$ & $34.9$ & $40.8$
& Max. Sinus Abn. & $0.1$ & $0.0$ & $0.1$ \\

Metallic Rest. & $34.9$ & $51.3$ & $35.8$
& & & & \\
\bottomrule
\end{tabular}%
}
\end{table*}

\subsection{Annotation Reproducibility}
\label{sec:annotation-reproducibility}
\TMItabref{tab:cross-annotator-validation} reports agreement on $900$ case--attribute decisions across $100$ OPGs. The $711$ agreements yield $79.0\%$ raw agreement ($95\%$ CI: $76.0\%$--$82.0\%$), pooled Cohen's $\kappa=0.558$ ($95\%$ CI: $0.501$--$0.613$), and Jaccard similarity $0.572$. Root-canal treatment shows $96\%$ agreement ($\kappa=0.920$, Jaccard $=0.922$), and tooth-supported bridge or crown findings show $93\%$ agreement ($\kappa=0.806$, Jaccard $=0.741$). Generalized crestal bone loss is less reproducible in the mandible ($64\%$, $\kappa=0.198$) and maxilla ($69\%$, $\kappa=0.271$). Repeat review identifies fewer bone-loss findings but more periapical lesions and restorations, suggesting attribute-specific interpretation thresholds. The binary review measures reproducibility, not clinical superiority of either annotation set or an upper bound on model performance.

\section{Experiments}
\subsection{Experimental Setup}
\label{sec:experimental-setup}
\paragraph{Training and Benchmark Splits}
All models are evaluated on the same $505$-case in-house benchmark, held out exclusively for final evaluation. Supervised fine-tuning uses all $9{,}049$ public-source cases. The two subsets are strictly disjoint, testing cross-source generalization.

\paragraph{Compared Models}
The evaluation covers proprietary general-domain models, open-source general-domain models, and medical-domain models. Proprietary models use official APIs with fixed prompts and deterministic decoding where supported. The complete model inventory and references are provided in \TMItabref{tab:nlg_raw_results} and \TMItabref{tab:attr_detailed_results}. OralGPT-Captioning-4B-Base~\cite{Fan2026OralGPTPlus} is the principal dental-specific baseline. Qwen3-VL-4B-Instruct~\cite{Qwen2025Qwen3VL} and Qwen3.5-4B-Instruct~\cite{Qwen2026Qwen35} are additionally fine-tuned on \textsc{PanDent} and evaluated with the same metrics as their zero-shot counterparts.

\begin{table*}[t!]
\centering
\small
    \renewcommand{\arraystretch}{0.95}
\setlength{\tabcolsep}{14.8pt}

\caption{Textual agreement on $505$ in-house OPGs. Entries: raw scores; subscripts: normalized-minus-raw changes (green/red: positive/negative). Average: unweighted mean of six metrics. Column-wise best/second-best: red/blue.}

\label{tab:nlg_raw_results}
\resizebox{\textwidth}{!}{%
\begin{tabular}{l r c c c c c c c}
\toprule
\textbf{\textsc{Method}} & \textbf{\textsc{Size (B)}} & \textbf{\textsc{BLEU-1}} & \textbf{\textsc{BLEU-2}} & \textbf{\textsc{BLEU-3}} & \textbf{\textsc{BLEU-4}} & \textbf{\textsc{METEOR}} & \textbf{\textsc{ROUGE-L}} & \textbf{\textsc{Average}} \\
\midrule
\rowcolor{gray!18}
\multicolumn{9}{l}{\textbf{General-Domain Closed-Source MLLMs}} \\

GPT-6.1 Sol~\cite{openai2026gpt61solsystemcard} & \nasize & \posdelta{0.411}{0.063} & \posdelta{0.232}{0.233} & \posdelta{0.131}{0.303} & \posdelta{0.063}{0.353} & \posdelta{0.189}{0.112} & \posdelta{0.225}{0.550} & \posdelta{0.209}{0.269} \\
GPT-6 Astra~\cite{openai2026gpt6astrasystemcard} & \nasize & \posdelta{0.413}{0.081} & \posdelta{0.248}{0.232} & \posdelta{0.141}{0.312} & \posdelta{0.066}{0.356} & \posdelta{0.196}{0.105} & \posdelta{0.224}{0.545} & \posdelta{0.215}{0.272} \\
GPT-6 Sol~\cite{openai2026gpt6solsystemcard} & \nasize & \posdelta{0.402}{0.079} & \posdelta{0.229}{0.229} & \posdelta{0.129}{0.296} & \posdelta{0.058}{0.346} & \posdelta{0.190}{0.110} & \posdelta{0.209}{0.535} & \posdelta{0.203}{0.266} \\

GPT-5.6 Luna~\cite{openai2026gpt56systemcard}
& \nasize
& \posdelta{0.381}{0.075}
& \posdelta{0.206}{0.226}
& \posdelta{0.107}{0.298}
& \posdelta{0.044}{0.337}
& \posdelta{0.188}{0.108}
& \posdelta{0.193}{0.521}
& \posdelta{0.187}{0.261} \\

GPT-5.6 Sol~\cite{openai2026gpt56systemcard}
& \nasize
& \posdelta{0.383}{0.079}
& \posdelta{0.215}{0.224}
& \posdelta{0.118}{0.311}
& \posdelta{0.053}{0.344}
& \posdelta{0.192}{0.100}
& \posdelta{0.189}{0.525}
& \posdelta{0.192}{0.264} \\

GPT-5.6 Terra~\cite{openai2026gpt56systemcard}
& \nasize
& \posdelta{0.399}{0.068}
& \posdelta{0.227}{0.243}
& \posdelta{0.127}{0.295}
& \posdelta{0.048}{0.352}
& \posdelta{0.192}{0.106}
& \posdelta{0.191}{0.537}
& \posdelta{0.197}{0.267} \\

GPT-5.4~\cite{openai2026gpt54thinking} & \nasize & \posdelta{0.366}{0.066} & \posdelta{0.206}{0.222} & \posdelta{0.116}{0.297} & \posdelta{0.033}{0.332} & \posdelta{0.177}{0.116} & \posdelta{0.186}{0.525} & \posdelta{0.181}{0.260} \\
GPT-5.1~\cite{openai2025gpt51systemcard} & \nasize & \posdelta{0.339}{0.057} & \posdelta{0.180}{0.181} & \posdelta{0.085}{0.251} & \posdelta{0.039}{0.270} & \posdelta{0.165}{0.075} & \posdelta{0.163}{0.395} & \posdelta{0.162}{0.205} \\
GPT-4.1~\cite{openai2025gpt41modeldocumentation} & \nasize & \posdelta{0.331}{0.065} & \posdelta{0.163}{0.201} & \posdelta{0.083}{0.253} & \posdelta{0.025}{0.300} & \posdelta{0.142}{0.075} & \posdelta{0.158}{0.366} & \posdelta{0.150}{0.210} \\

Gemini 4 Argon~\cite{googledeepmind2026gemini4argon} & \nasize & \posdelta{0.363}{0.086} & \posdelta{0.197}{0.205} & \posdelta{0.103}{0.292} & \posdelta{0.046}{0.334} & \posdelta{0.180}{0.104} & \posdelta{0.187}{0.499} & \posdelta{0.179}{0.253} \\
Gemini 3.8 Flash~\cite{googledeepmind2026gemini38flash} & \nasize & \posdelta{0.336}{0.060} & \posdelta{0.167}{0.216} & \posdelta{0.097}{0.268} & \posdelta{0.041}{0.306} & \posdelta{0.147}{0.092} & \posdelta{0.168}{0.465} & \posdelta{0.159}{0.235} \\
Gemini 3.7 Flash~\cite{googledeepmind2026gemini37flash} & \nasize & \posdelta{0.325}{0.063} & \posdelta{0.159}{0.188} & \posdelta{0.074}{0.267} & \posdelta{0.029}{0.303} & \posdelta{0.146}{0.091} & \posdelta{0.166}{0.454} & \posdelta{0.150}{0.228} \\
Gemini 3.6 Flash~\cite{googledeepmind2026gemini36flash} & \nasize & \posdelta{0.302}{0.058} & \posdelta{0.155}{0.188} & \posdelta{0.068}{0.271} & \posdelta{0.028}{0.292} & \posdelta{0.135}{0.086} & \posdelta{0.164}{0.431} & \posdelta{0.142}{0.221} \\
Gemini 3.5 Flash~\cite{googledeepmind2026gemini35flash} & \nasize & \posdelta{0.295}{0.061} & \posdelta{0.140}{0.180} & \posdelta{0.063}{0.249} & \posdelta{0.021}{0.288} & \posdelta{0.133}{0.089} & \posdelta{0.148}{0.416} & \posdelta{0.133}{0.214} \\
Gemini 3 Pro Preview~\cite{google2025gemini3pro} & \nasize & \negdelta{0.202}{0.196} & \negdelta{0.080}{0.076} & \negdelta{0.035}{0.031} & \negdelta{0.017}{0.013} & \negdelta{0.093}{0.040} & \posdelta{0.105}{0.008} & \negdelta{0.089}{0.058} \\
Gemini 3 Flash Preview~\cite{googledeepmind2025gemini3flash} & \nasize & \posdelta{0.274}{0.057} & \posdelta{0.125}{0.180} & \posdelta{0.055}{0.245} & \posdelta{0.017}{0.280} & \posdelta{0.133}{0.079} & \posdelta{0.138}{0.392} & \posdelta{0.124}{0.206} \\
Grok 4.7~\cite{xai2026grok47modelcard} & \nasize & \posdelta{0.355}{0.072} & \posdelta{0.179}{0.209} & \posdelta{0.091}{0.273} & \posdelta{0.030}{0.322} & \posdelta{0.163}{0.095} & \posdelta{0.189}{0.479} & \posdelta{0.168}{0.242} \\
Grok 4.6~\cite{xai2026grok46modelcard} & \nasize & \posdelta{0.324}{0.053} & \posdelta{0.170}{0.200} & \posdelta{0.076}{0.268} & \posdelta{0.028}{0.312} & \posdelta{0.165}{0.100} & \posdelta{0.175}{0.457} & \posdelta{0.156}{0.232} \\
Grok 4.3~\cite{xai2026grok43modeldocumentation} & \nasize & \posdelta{0.316}{0.061} & \posdelta{0.167}{0.199} & \posdelta{0.075}{0.269} & \posdelta{0.024}{0.285} & \posdelta{0.136}{0.096} & \posdelta{0.164}{0.420} & \posdelta{0.147}{0.222} \\
Grok 4.20 Reasoning~\cite{xai2026grok420systemcard} & \nasize & \posdelta{0.305}{0.072} & \posdelta{0.149}{0.189} & \posdelta{0.074}{0.252} & \posdelta{0.023}{0.290} & \posdelta{0.138}{0.084} & \posdelta{0.145}{0.376} & \posdelta{0.139}{0.211} \\
Claude Opus 5~\cite{anthropic2026claudeopus5}
& \nasize
& \posdelta{0.380}{0.074}
& \posdelta{0.194}{0.223}
& \posdelta{0.109}{0.288}
& \posdelta{0.049}{0.345}
& \posdelta{0.173}{0.112}
& \posdelta{0.198}{0.523}
& \posdelta{0.184}{0.261} \\

Claude Opus 4.7~\cite{anthropic2026claudeopus47} & \nasize & \posdelta{0.354}{0.077} & \posdelta{0.197}{0.208} & \posdelta{0.109}{0.299} & \posdelta{0.043}{0.326} & \posdelta{0.160}{0.103} & \posdelta{0.188}{0.524} & \posdelta{0.175}{0.256} \\
Claude Opus 4.1~\cite{anthropic2025claudeopus41} & \nasize & \posdelta{0.312}{0.056} & \posdelta{0.151}{0.191} & \posdelta{0.071}{0.257} & \posdelta{0.025}{0.288} & \posdelta{0.149}{0.082} & \posdelta{0.157}{0.369} & \posdelta{0.144}{0.207} \\
Claude Sonnet 4~\cite{anthropic2025claude4} & \nasize & \posdelta{0.300}{0.064} & \posdelta{0.135}{0.196} & \posdelta{0.054}{0.257} & \posdelta{0.023}{0.301} & \posdelta{0.147}{0.080} & \posdelta{0.159}{0.382} & \posdelta{0.136}{0.213} \\

\midrule
\rowcolor{gray!18}
\multicolumn{9}{l}{\textbf{General-Domain Open-Source MLLMs}} \\
Qwen3-VL-235B-A22B-Instruct~\cite{Qwen2025Qwen3VL} & 235 & \posdelta{0.286}{0.101} & \posdelta{0.132}{0.208} & \posdelta{0.058}{0.263} & \posdelta{0.028}{0.271} & \posdelta{0.183}{0.057} & \posdelta{0.170}{0.349} & \posdelta{0.143}{0.208} \\
Qwen3-VL-4B-Instruct~\cite{Qwen2025Qwen3VL} & 4 & \posdelta{0.219}{0.141} & \posdelta{0.093}{0.247} & \posdelta{0.037}{0.303} & \posdelta{0.016}{0.326} & \posdelta{0.150}{0.082} & \posdelta{0.144}{0.401} & \posdelta{0.110}{0.250} \\
Qwen3-VL-8B-Instruct~\cite{Qwen2025Qwen3VL} & 8 & \posdelta{0.228}{0.133} & \posdelta{0.100}{0.236} & \posdelta{0.038}{0.309} & \posdelta{0.019}{0.312} & \posdelta{0.150}{0.073} & \posdelta{0.159}{0.413} & \posdelta{0.116}{0.246} \\
Qwen3.5-4B-Instruct~\cite{Qwen2026Qwen35} & 4 & \posdelta{0.216}{0.126} & \posdelta{0.101}{0.257} & \posdelta{0.055}{0.301} & \posdelta{0.017}{0.323} & \posdelta{0.145}{0.083} & \posdelta{0.146}{0.394} & \posdelta{0.113}{0.247} \\
Qwen3.5-9B-Instruct~\cite{Qwen2026Qwen35} & 9 & \posdelta{0.238}{0.123} & \posdelta{0.106}{0.240} & \posdelta{0.043}{0.298} & \posdelta{0.018}{0.315} & \posdelta{0.163}{0.085} & \posdelta{0.154}{0.396} & \posdelta{0.120}{0.243} \\
Qwen2.5-VL-72B-Instruct~\cite{Qwen2025Qwen25VL} & 72 & \posdelta{0.252}{0.108} & \posdelta{0.113}{0.228} & \posdelta{0.055}{0.280} & \posdelta{0.041}{0.295} & \posdelta{0.177}{0.047} & \posdelta{0.150}{0.364} & \posdelta{0.131}{0.220} \\
InternVL3-78B~\cite{Zhu2025InternVL3} & 78 & \posdelta{0.260}{0.106} & \posdelta{0.117}{0.210} & \posdelta{0.063}{0.259} & \posdelta{0.023}{0.284} & \posdelta{0.169}{0.053} & \posdelta{0.168}{0.371} & \posdelta{0.133}{0.214} \\
MiniCPM-V-4\_5~\cite{Yao2025MiniCPMV45} & 8 & \posdelta{0.194}{0.225} & \posdelta{0.084}{0.307} & \posdelta{0.033}{0.334} & \posdelta{0.014}{0.349} & \posdelta{0.133}{0.110} & \posdelta{0.137}{0.421} & \posdelta{0.099}{0.291} \\
Kimi-VL-A3B-Thinking-2506~\cite{Moonshot2025KimiVL} & 16 & \posdelta{0.254}{0.105} & \posdelta{0.113}{0.226} & \posdelta{0.046}{0.284} & \posdelta{0.022}{0.291} & \posdelta{0.162}{0.066} & \posdelta{0.152}{0.386} & \posdelta{0.125}{0.226} \\
GLM-4.1V-9B-Thinking~\cite{Zhipu2025GLM41V} & 9 & \posdelta{0.204}{0.236} & \posdelta{0.088}{0.334} & \posdelta{0.028}{0.372} & \posdelta{0.012}{0.378} & \posdelta{0.145}{0.114} & \posdelta{0.131}{0.452} & \posdelta{0.101}{0.314} \\
DeepSeek-VL2~\cite{DeepSeekAI2024VL2} & 27 & \posdelta{0.238}{0.103} & \posdelta{0.111}{0.225} & \posdelta{0.046}{0.271} & \posdelta{0.013}{0.297} & \posdelta{0.154}{0.061} & \posdelta{0.146}{0.376} & \posdelta{0.118}{0.222} \\
Pixtral-Large-Instruct-2411~\cite{MistralAI2024Pixtral} & 124 & \posdelta{0.272}{0.106} & \posdelta{0.122}{0.211} & \posdelta{0.057}{0.263} & \posdelta{0.025}{0.279} & \posdelta{0.179}{0.052} & \posdelta{0.164}{0.358} & \posdelta{0.137}{0.212} \\
Gemma-3-27B-IT~\cite{GemmaTeam2025Gemma3} & 27 & \posdelta{0.228}{0.128} & \posdelta{0.112}{0.232} & \posdelta{0.044}{0.274} & \posdelta{0.016}{0.294} & \posdelta{0.148}{0.076} & \posdelta{0.136}{0.377} & \posdelta{0.114}{0.230} \\
Llama-3.2-90B-Vision-Instruct~\cite{Dubey2024Llama3} & 90 & \posdelta{0.054}{0.401} & \posdelta{0.024}{0.412} & \posdelta{0.001}{0.410} & \posdelta{0.000}{0.392} & \posdelta{0.017}{0.237} & \posdelta{0.049}{0.537} & \posdelta{0.024}{0.398} \\
\midrule
\rowcolor{gray!18}
\multicolumn{9}{l}{\textbf{Medical-Domain MLLMs}} \\
MedGemma-27B-IT~\cite{MedGemmaTeam2025MedGemma} & 27 & \posdelta{0.228}{0.110} & \posdelta{0.104}{0.224} & \posdelta{0.045}{0.271} & \posdelta{0.016}{0.291} & \posdelta{0.136}{0.069} & \posdelta{0.132}{0.378} & \posdelta{0.110}{0.224} \\
MedGemma-4B-IT~\cite{MedGemmaTeam2025MedGemma} & 4 & \posdelta{0.218}{0.126} & \posdelta{0.096}{0.228} & \posdelta{0.042}{0.270} & \posdelta{0.019}{0.283} & \posdelta{0.128}{0.093} & \posdelta{0.153}{0.348} & \posdelta{0.109}{0.225} \\
Lingshu-32B~\cite{Zhang2025Lingshu} & 32 & \posdelta{0.188}{0.122} & \posdelta{0.075}{0.241} & \posdelta{0.034}{0.291} & \posdelta{0.014}{0.302} & \posdelta{0.111}{0.087} & \posdelta{0.121}{0.400} & \posdelta{0.091}{0.241} \\
PulseMind-72B~\cite{Xu2026PulseMind} & 72 & \posdelta{0.201}{0.124} & \posdelta{0.087}{0.241} & \posdelta{0.037}{0.278} & \posdelta{0.015}{0.298} & \posdelta{0.118}{0.072} & \posdelta{0.121}{0.394} & \posdelta{0.097}{0.235} \\
MAIRA-2~\cite{Jin2024MAIRA2} & 7 & \posdelta{0.030}{0.433} & \posdelta{0.002}{0.426} & \posdelta{0.006}{0.404} & \posdelta{0.003}{0.394} & \posdelta{0.034}{0.227} & \posdelta{0.097}{0.486} & \posdelta{0.029}{0.395} \\
        CheXagent-8B~\cite{Chen2024CheXagent} & 8 & \posdelta{0.162}{0.137} & \posdelta{0.067}{0.262} & \posdelta{0.027}{0.301} & \posdelta{0.011}{0.315} & \posdelta{0.098}{0.092} & \posdelta{0.104}{0.420} & \posdelta{0.078}{0.255} \\
        Libra-v1.0-3B~\cite{Zhang2025Libra} & 3 & \posdelta{0.149}{0.144} & \posdelta{0.060}{0.273} & \posdelta{0.024}{0.307} & \posdelta{0.009}{0.317} & \posdelta{0.087}{0.096} & \posdelta{0.096}{0.433} & \posdelta{0.071}{0.262} \\
    MVL-RRG-1.0~\cite{SNUH2026MVLRRG} & 8 & \posdelta{0.080}{0.367} & \posdelta{0.037}{0.386} & \posdelta{0.018}{0.390} & \posdelta{0.009}{0.379} & \posdelta{0.053}{0.211} & \posdelta{0.118}{0.455} & \posdelta{0.053}{0.365} \\
OralGPT-Captioning-4B-Base~\cite{Fan2026OralGPTPlus} & 4 & \posdelta{0.221}{0.240} & \posdelta{0.110}{0.324} & \posdelta{0.049}{0.374} & \posdelta{0.021}{0.390} & \posdelta{0.117}{0.174} & \posdelta{0.125}{0.604} & \posdelta{0.107}{0.351} \\
\midrule
\rowcolor{gray!18}
\multicolumn{9}{l}{\textbf{Finetuned Baselines}} \\
Qwen3-VL-4B-Instruct (Ours) & 4 & \second{\posdelta{0.862}{0.000}} & \second{\posdelta{0.825}{0.000}} & \second{\posdelta{0.792}{0.000}} & \second{\posdelta{0.767}{0.000}} & \second{\posdelta{0.482}{0.000}} & \second{\posdelta{0.829}{0.000}} & \second{\posdelta{0.760}{0.000}} \\
Qwen3.5-4B-Instruct (Ours) & 4 & \best{\posdelta{0.871}{0.000}} & \best{\posdelta{0.840}{0.000}} & \best{\posdelta{0.808}{0.000}} & \best{\posdelta{0.785}{0.000}} & \best{\posdelta{0.494}{0.000}} & \best{\posdelta{0.850}{0.000}} & \best{\posdelta{0.775}{0.000}} \\
\bottomrule
\end{tabular}%
}

\end{table*}

\begin{table*}[t!]
\centering
\scriptsize
\renewcommand{\arraystretch}{0.88}
\setlength{\tabcolsep}{8.2pt}
\caption{Report-derived attribute exact-match accuracy on $505$ in-house OPGs. Correctness requires matching the binary status for each regional finding or the complete FDI set for each tooth-level finding. Average gives equal weight to all attributes.}
\label{tab:attr_detailed_results}
\resizebox{\textwidth}{!}{%
\begin{tabular}{l r c c c c c c c c c c}
\toprule
\multirow{2}{*}{\textbf{\textsc{Method}}} & \multirow{2}{*}{\textbf{\textsc{Size (B)}}} & \multicolumn{2}{c}{\textbf{\textsc{Binary Regional-State}}} & \multicolumn{7}{c}{\textbf{\textsc{Tooth-Index Grounding}}} & \multirow{2}{*}{\textbf{\textsc{Average}}} \\
\cmidrule(lr){3-4}\cmidrule(lr){5-11}
 & & \textbf{\textsc{GCBL-M}} & \textbf{\textsc{GCBL-X}} & \textbf{\textsc{Post}} & \textbf{\textsc{Peri.}} & \textbf{\textsc{RCT}} & \textbf{\textsc{MR}} & \textbf{\textsc{Caries}} & \textbf{\textsc{NMR}} & \textbf{\textsc{Crown}} &  \\
\midrule
\rowcolor{gray!18}
\multicolumn{12}{l}{\textbf{General-Domain Closed-Source MLLMs}} \\

GPT-6.1 Sol~\cite{openai2026gpt61solsystemcard} & \nasize & 0.450 & 0.444 & 0.453 & 0.465 & 0.455 & 0.434 & 0.459 & 0.432 & 0.459 & 0.450 \\
GPT-6 Astra~\cite{openai2026gpt6astrasystemcard} & \nasize & 0.465 & 0.455 & 0.473 & 0.465 & 0.446 & 0.451 & 0.448 & 0.444 & 0.463 & 0.457 \\
GPT-6 Sol~\cite{openai2026gpt6solsystemcard} & \nasize & 0.463 & 0.436 & 0.463 & 0.451 & 0.451 & 0.428 & 0.430 & 0.442 & 0.438 & 0.445 \\

GPT-5.6 Luna~\cite{openai2026gpt56systemcard}
& \nasize
& 0.440 & 0.406
& 0.440 & 0.434 & 0.412 & 0.440
& 0.434 & 0.406 & 0.434
& 0.427 \\

GPT-5.6 Sol~\cite{openai2026gpt56systemcard}
& \nasize
& 0.446 & 0.426
& 0.438 & 0.444 & 0.432 & 0.424
& 0.422 & 0.422 & 0.446
& 0.433 \\

GPT-5.6 Terra~\cite{openai2026gpt56systemcard}
& \nasize
& 0.436 & 0.448
& 0.424 & 0.432 & 0.436 & 0.428
& 0.457 & 0.444 & 0.444
& 0.439 \\

GPT-5.4~\cite{openai2026gpt54thinking} & \nasize & 0.430 & 0.428 & 0.420 & 0.404 & 0.410 & 0.408 & 0.408 & 0.416 & 0.436 & 0.418 \\
GPT-5.1~\cite{openai2025gpt51systemcard} & \nasize & 0.398 & 0.408 & 0.390 & 0.412 & 0.414 & 0.400 & 0.422 & 0.394 & 0.410 & 0.405 \\
GPT-4.1~\cite{openai2025gpt41modeldocumentation} & \nasize & 0.358 & 0.376 & 0.362 & 0.382 & 0.394 & 0.358 & 0.370 & 0.376 & 0.382 & 0.373 \\

Gemini 4 Argon~\cite{googledeepmind2026gemini4argon} & \nasize & 0.434 & 0.416 & 0.438 & 0.444 & 0.424 & 0.422 & 0.438 & 0.418 & 0.451 & 0.431 \\
Gemini 3.8 Flash~\cite{googledeepmind2026gemini38flash} & \nasize & 0.414 & 0.400 & 0.416 & 0.418 & 0.420 & 0.422 & 0.404 & 0.392 & 0.412 & 0.411 \\
Gemini 3.7 Flash~\cite{googledeepmind2026gemini37flash} & \nasize & 0.378 & 0.410 & 0.410 & 0.404 & 0.380 & 0.404 & 0.394 & 0.380 & 0.404 & 0.396 \\
Gemini 3.6 Flash~\cite{googledeepmind2026gemini36flash} & \nasize & 0.366 & 0.404 & 0.386 & 0.410 & 0.382 & 0.382 & 0.382 & 0.366 & 0.386 & 0.385 \\
Gemini 3.5 Flash~\cite{googledeepmind2026gemini35flash} & \nasize & 0.370 & 0.358 & 0.366 & 0.378 & 0.378 & 0.372 & 0.370 & 0.358 & 0.392 & 0.372 \\
Gemini 3 Pro Preview~\cite{google2025gemini3pro} & \nasize & 0.394 & 0.410 & 0.398 & 0.410 & 0.404 & 0.384 & 0.392 & 0.396 & 0.394 & 0.398 \\
Gemini 3 Flash Preview~\cite{googledeepmind2025gemini3flash} & \nasize & 0.360 & 0.366 & 0.356 & 0.356 & 0.360 & 0.360 & 0.376 & 0.341 & 0.356 & 0.359 \\
Grok 4.7~\cite{xai2026grok47modelcard} & \nasize & 0.408 & 0.402 & 0.426 & 0.414 & 0.410 & 0.422 & 0.436 & 0.402 & 0.430 & 0.417 \\
Grok 4.6~\cite{xai2026grok46modelcard} & \nasize & 0.394 & 0.384 & 0.424 & 0.404 & 0.400 & 0.412 & 0.408 & 0.404 & 0.400 & 0.403 \\
Grok 4.3~\cite{xai2026grok43modeldocumentation} & \nasize & 0.398 & 0.388 & 0.418 & 0.382 & 0.388 & 0.380 & 0.392 & 0.402 & 0.386 & 0.393 \\
Grok 4.20 Reasoning~\cite{xai2026grok420systemcard} & \nasize & 0.372 & 0.376 & 0.394 & 0.400 & 0.384 & 0.390 & 0.382 & 0.362 & 0.368 & 0.381 \\

Claude Opus 5~\cite{anthropic2026claudeopus5}
& \nasize
& 0.434 & 0.424
& 0.428 & 0.428 & 0.412 & 0.428
& 0.434 & 0.408 & 0.416
& 0.423 \\

Claude Opus 4.7~\cite{anthropic2026claudeopus47} & \nasize & 0.422 & 0.420 & 0.432 & 0.406 & 0.406 & 0.412 & 0.394 & 0.400 & 0.412 & 0.411 \\
Claude Opus 4.1~\cite{anthropic2025claudeopus41} & \nasize & 0.386 & 0.368 & 0.404 & 0.412 & 0.402 & 0.380 & 0.388 & 0.372 & 0.386 & 0.389 \\
Claude Sonnet 4~\cite{anthropic2025claude4} & \nasize & 0.352 & 0.380 & 0.362 & 0.372 & 0.378 & 0.366 & 0.350 & 0.347 & 0.364 & 0.364 \\

\midrule
\rowcolor{gray!18}
\multicolumn{12}{l}{\textbf{General-Domain Open-Source MLLMs}} \\
Qwen3-VL-235B-A22B-Instruct~\cite{Qwen2025Qwen3VL} & 235 & 0.321 & 0.335 & 0.356 & 0.366 & 0.343 & 0.331 & 0.333 & 0.345 & 0.349 & 0.342 \\
Qwen3-VL-4B-Instruct~\cite{Qwen2025Qwen3VL} & 4 & 0.244 & 0.265 & 0.271 & 0.275 & 0.279 & 0.265 & 0.261 & 0.271 & 0.291 & 0.269 \\
Qwen3-VL-8B-Instruct~\cite{Qwen2025Qwen3VL} & 8 & 0.273 & 0.293 & 0.295 & 0.307 & 0.285 & 0.275 & 0.279 & 0.293 & 0.277 & 0.286 \\
Qwen3.5-4B-Instruct~\cite{Qwen2026Qwen35} & 4 & 0.271 & 0.271 & 0.283 & 0.281 & 0.273 & 0.271 & 0.279 & 0.273 & 0.283 & 0.276 \\
Qwen3.5-9B-Instruct~\cite{Qwen2026Qwen35} & 9 & 0.271 & 0.277 & 0.305 & 0.293 & 0.297 & 0.305 & 0.291 & 0.295 & 0.321 & 0.295 \\
Qwen2.5-VL-72B-Instruct~\cite{Qwen2025Qwen25VL} & 72 & 0.295 & 0.307 & 0.341 & 0.309 & 0.299 & 0.311 & 0.327 & 0.311 & 0.319 & 0.313 \\
InternVL3-78B~\cite{Zhu2025InternVL3} & 78 & 0.329 & 0.305 & 0.321 & 0.337 & 0.311 & 0.323 & 0.309 & 0.321 & 0.329 & 0.320 \\
MiniCPM-V-4\_5~\cite{Yao2025MiniCPMV45} & 8 & 0.255 & 0.269 & 0.287 & 0.291 & 0.267 & 0.279 & 0.281 & 0.261 & 0.271 & 0.274 \\
Kimi-VL-A3B-Thinking-2506~\cite{Moonshot2025KimiVL} & 16 & 0.251 & 0.267 & 0.279 & 0.267 & 0.253 & 0.261 & 0.269 & 0.238 & 0.265 & 0.261 \\
GLM-4.1V-9B-Thinking~\cite{Zhipu2025GLM41V} & 9 & 0.287 & 0.275 & 0.269 & 0.291 & 0.263 & 0.271 & 0.281 & 0.283 & 0.305 & 0.281 \\
DeepSeek-VL2~\cite{DeepSeekAI2024VL2} & 27 & 0.269 & 0.293 & 0.293 & 0.311 & 0.275 & 0.293 & 0.307 & 0.285 & 0.313 & 0.293 \\
Pixtral-Large-Instruct-2411~\cite{MistralAI2024Pixtral} & 124 & 0.311 & 0.339 & 0.337 & 0.343 & 0.335 & 0.345 & 0.323 & 0.343 & 0.333 & 0.334 \\
Gemma-3-27B-IT~\cite{GemmaTeam2025Gemma3} & 27 & 0.287 & 0.281 & 0.313 & 0.281 & 0.295 & 0.295 & 0.285 & 0.299 & 0.303 & 0.293 \\
Llama-3.2-90B-Vision-Instruct~\cite{Dubey2024Llama3} & 90 & 0.339 & 0.317 & 0.317 & 0.331 & 0.337 & 0.327 & 0.331 & 0.317 & 0.319 & 0.326 \\
\midrule
\rowcolor{gray!18}
\multicolumn{12}{l}{\textbf{Medical-Domain MLLMs}} \\
MedGemma-27B-IT~\cite{MedGemmaTeam2025MedGemma} & 27 & 0.117 & 0.105 & 0.101 & 0.105 & 0.107 & 0.091 & 0.117 & 0.111 & 0.113 & 0.107 \\
MedGemma-4B-IT~\cite{MedGemmaTeam2025MedGemma} & 4 & 0.107 & 0.109 & 0.093 & 0.101 & 0.091 & 0.093 & 0.111 & 0.097 & 0.101 & 0.100 \\
Lingshu-32B~\cite{Zhang2025Lingshu} & 32 & 0.103 & 0.121 & 0.109 & 0.123 & 0.101 & 0.111 & 0.109 & 0.119 & 0.113 & 0.112 \\
PulseMind-72B~\cite{Xu2026PulseMind} & 72 & 0.103 & 0.121 & 0.119 & 0.115 & 0.125 & 0.109 & 0.121 & 0.111 & 0.135 & 0.117 \\
MAIRA-2~\cite{Jin2024MAIRA2} & 7 & 0.087 & 0.087 & 0.107 & 0.113 & 0.081 & 0.079 & 0.101 & 0.103 & 0.097 & 0.095 \\
CheXagent-8B~\cite{Chen2024CheXagent} & 8 & 0.105 & 0.115 & 0.125 & 0.095 & 0.109 & 0.095 & 0.093 & 0.087 & 0.111 & 0.104 \\
Libra-v1.0-3B~\cite{Zhang2025Libra} & 3 & 0.089 & 0.101 & 0.095 & 0.089 & 0.077 & 0.087 & 0.097 & 0.101 & 0.085 & 0.091 \\
MVL-RRG-1.0~\cite{SNUH2026MVLRRG} & 8 & 0.085 & 0.109 & 0.109 & 0.089 & 0.107 & 0.093 & 0.087 & 0.089 & 0.101 & 0.097 \\
OralGPT-Captioning-4B-Base~\cite{Fan2026OralGPTPlus} & 4 & 0.515 & 0.459 & 0.743 & 0.471 & 0.366 & 0.248 & 0.024 & 0.212 & 0.568 & 0.401 \\
\midrule
\rowcolor{gray!18}
\multicolumn{12}{l}{\textbf{Finetuned Baselines}} \\
Qwen3-VL-4B-Instruct (Ours) & 4 & \second{0.642} & \second{0.632} & \second{0.758} & \second{0.618} & \second{0.602} & \second{0.570} & \second{0.539} & \second{0.556} & \second{0.671} & \second{0.621} \\
Qwen3.5-4B-Instruct (Ours) & 4 & \best{0.646} & \best{0.661} & \best{0.776} & \best{0.630} & \best{0.630} & \best{0.586} & \best{0.554} & \best{0.566} & \best{0.691} & \best{0.638} \\
\bottomrule
\end{tabular}%
}

\noindent\parbox{\textwidth}{%
\scriptsize
\textbf{\textsc{Attribute Abbreviations.}}

\begin{tabular}[t]{@{}p{0.48\textwidth}p{0.48\textwidth}@{}}
\textbf{\textsc{GCBL-M/X}}: mandibular/maxillary generalized crestal bone loss; \textbf{\textsc{Post}}: post and core; \textbf{\textsc{Peri.}}: periapical lesion; &
\textbf{\textsc{RCT}}: root-canal treatment; \textbf{\textsc{MR}}: metallic restoration; \textbf{\textsc{NMR}}: non-metallic restoration; \textbf{\textsc{Crown}}: tooth supported crown.
\end{tabular}
}

\end{table*}

\begin{table*}[t!]
    \centering
    \caption{In-house case: OPG and reference findings (left); reference/model reports and raw-to-normalized pairs (right); backgrounds align clinical attributes; bold underlined green/red spans mark correct/incorrect or missing descriptions.}
    \label{tab:case_study}
    \scriptsize
    {\hfuzz=2pt
    \scalebox{0.935}{\begin{minipage}{1.0256\textwidth}
    \begin{tabularx}{\linewidth}{>{\centering\arraybackslash}p{0.30\linewidth}>{\raggedright\arraybackslash}X}
        \toprule
        \textbf{\textsc{OPG and Reference Findings}} &
        \textbf{\textsc{Reference Report and Qualitative Comparison}} \\
        \midrule
        \begin{minipage}[t]{\linewidth}
        \hrule height 0pt\relax
        
        \centering
        \includegraphics[width=\linewidth,keepaspectratio]{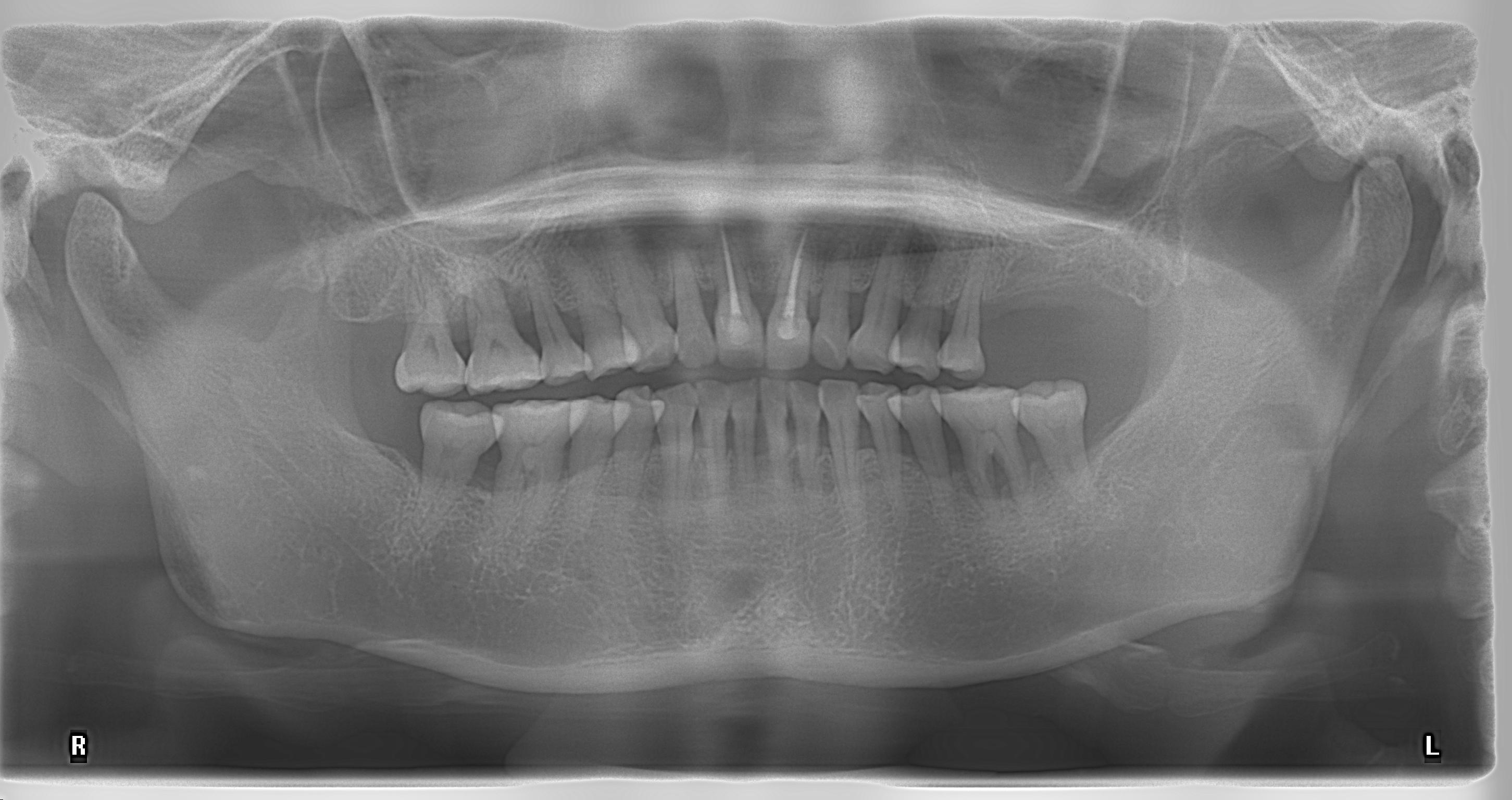}

        \parbox{\linewidth}{\raggedright\textbf{\textsc{Prompt:}} Generate a free-text radiology report according to the input radiograph.}

        \caseattrbox{\raggedright
        \casemissing{Missing}: \#18, \#26, \#27, \#28, \#38, \#48;\par
        \casetreat{Non-metallic restoration}: \#11, \#21;\par
        \casetreat{Root-canal treatment}: \#11, \#21;\par
        \casebone{Generalized crestal bone loss}: maxilla and mandible.}
        \end{minipage}
        &
        \begin{minipage}[t]{\linewidth}
        \hrule height 0pt\relax
        
        \fontsize{6.78}{6.86}\selectfont
        \raggedright

        \noindent\textbf{Reference report:}
        The panoramic radiograph shows \casemissing{absence of teeth} \#18, \#26, \#27, \#28, \#38, and \#48.
        \casetreat{Non-metal restorations} are present on teeth \#11 and \#21.
        \casetreat{Endodontic treatment} is evident in teeth \#11 and \#21.
        There is evidence of \casebone{generalized crestal bone loss} in the maxillary arch and \casebone{generalized crestal bone loss} in the mandibular arch.

        \reportsep
        \noindent\textbf{\textsc{GPT-5.4}:}
        Panoramic radiograph demonstrates a \casemissing{partially edentulous dentition} with \casegood{multiple missing posterior teeth bilaterally}, but \casebad{no explicit FDI tooth indices}.
        Mild \casebone{generalized horizontal alveolar bone loss} is \casegood{present}.
        \casetreat{Endodontic treatment changes} are seen in the \casegood{maxillary central incisors}, while the \casetreat{non-metal restorations} on \casebad{\#11 and \#21 are not mentioned}.
        No obvious focal periapical radiolucency, expansile osseous lesion, or acute fracture is identified; the \caseanat{visualized maxillary sinuses} are grossly clear.

        \reportsep
        \noindent\textbf{\textsc{Qwen3-VL-4B}:}
        \caserawnorm{
        \textbf{\textsc{Raw:}} This is a dental panoramic radiograph, also known as a panoramic X-ray. Key observations: \casebad{all 32 teeth visible}; roots and crowns are clearly delineated; the jawbone appears normal with no severe abnormality; ...}
        {\textbf{\textsc{Normalized:}} The output describes \casemissing{all 32 teeth visible} and a normal adult dentition, which is \casebad{incorrect} for the missing-tooth attribute. It also reports no large lesions, fractures, abscesses, significant caries, \casebad{visible implants or fillings}, and normal TMJs. It further states that the alveolar bone has \casebad{normal density and contour}, missing the \casebone{generalized crestal bone loss}. The \casetreat{restorations} and \casetreat{root-canal treatments} on \casebad{\#11 and \#21} are not recovered.}

        \reportsep
        \noindent\textbf{\textsc{OralGPT}:}
        \caserawnorm{
        \textbf{\textsc{Raw:}} \#\#\# Teeth-Specific Observations. General Condition: 25 teeth visualized; \casebad{1 missing tooth} \casebad{detected on the lower side}. Pathological Findings: \casebad{\#12, \#13, \#14, ...}, \casebad{\#45 suspicious} \casebad{for caries}; ...}
        {\textbf{\textsc{Normalized:}} The model produces a dental-style structured report, states that \casebad{25 teeth are visualized} with clear anatomical definition, and detects \casemissing{one missing tooth on the lower side}, but the finding is \casebad{under-specified} relative to six missing teeth. It additionally predicts \casebad{suspicious caries on \#12--\#17}, \casebad{\#21--\#27}, \casebad{\#31--\#37}, and \casebad{\#41--\#45} with re-evaluation suggestions, although the ground truth reports no caries. The \casetreat{restorations}, \casetreat{root-canal treatments}, and \casebone{generalized bone loss} are \casebad{not reliably recovered}.}

        \reportsep
        \noindent\modeldentbox{\textbf{\textsc{Fine-tuned Qwen3-VL-4B}:}
        The model reports \casemissing{absence of teeth} \casegood{\#18, \#28, \#38, and \#48}, but \casebad{omits \#26 and \#27}.
        It correctly recovers \casetreat{non-metal restorations} on \casegood{\#11 and \#21}, \casetreat{endodontic treatment} on \casegood{\#11 and \#21}, and \casebone{generalized crestal bone loss} in \casegood{both arches}.}
        
    \end{minipage} \\
        \bottomrule
    \end{tabularx}
    \end{minipage}}
    }
    
\end{table*}

\subsection{Evaluation Protocol}
\label{sec:evaluation_protocol}
Each generated report is evaluated at two complementary levels: \textit{report-level textual agreement} and \textit{attribute-level clinical diagnostic accuracy}. Textual agreement is assessed by comparing generated reports with their references. Clinical accuracy is assessed by extracting selected tooth-level and regional findings from the generated reports and comparing them with the corresponding structured annotations.

\paragraph{Report-Level Textual Agreement}
BLEU-$n$ ($n\in\{1,2,3,4\}$), METEOR, and ROUGE-L measure lexical overlap and sequence-level similarity between generated and reference reports. Higher scores indicate stronger textual agreement. Their unweighted arithmetic mean provides a compact model-level score.

\paragraph{Attribute-Level Clinical Diagnostic Accuracy}
Clinical findings corresponding to nine prioritized attributes are extracted from each generated report and mapped to the canonical annotation schema. For the seven tooth-index-aware attributes, extraction identifies the reported findings and their associated FDI tooth indices. For the two regional attributes, extraction identifies the reported abnormality status. The extracted findings are compared with the structured reference annotations to assess the clinical consistency of the generated report.
A tooth-index-aware prediction is correct only if its FDI tooth set exactly matches the reference; a regional prediction must match the binary abnormality status. Accuracy is reported per attribute and as the unweighted mean across all nine attributes:
\begin{equation}
\mathrm{Acc}_j=\frac{1}{N}\sum_{i=1}^{N}\mathbb{I}(\hat{z}_{i,j}=z_{i,j}).
\label{eq:attribute-accuracy}
\end{equation}
Here, $N=505$, $\hat{z}_{i,j}=E_j(\hat{R}_i)$, where $E_j$ denotes the extraction procedure for attribute $j$, $\hat{z}_{i,j}$ is the structured prediction derived from the generated report $\hat{R}_i$, and $z_{i,j}$ denotes the reference annotation for attribute $j$.

Attribute extraction prompts operate on the generated report and request one English sentence restricted to the target finding, excluding unrelated observations, markup, explanations, and recommendations. Tooth-index-aware prompts require valid two-digit permanent-tooth FDI indices and specify the quadrant mapping: upper right $11$--$18$, upper left $21$--$28$, lower left $31$--$38$, and lower right $41$--$48$. Binary regional prompts extract the reported presence or absence of abnormalities without tooth numbering. Explicitly negated findings are rendered using a fixed negative-sentence style.


\paragraph{Report Normalization}
A frozen Gemini 3.5 Flash~\cite{googledeepmind2026gemini35flash} rewrites each generated report into a concise clinician-guided schema, standardizing syntax, verbosity, formatting, and organization while preserving explicitly stated findings, tooth indices, diagnostic decisions, and regional details. The prompt prohibits adding unsupported content or completing unstated findings to match the style example. Ambiguous statements remain nonspecific, and headings, bullets, numbering, and analysis tags are removed. For the Gemini-based attribute extraction described above, clinicians audit a random sample of $200$ normalized outputs against their originals and identify no factual additions or clinically relevant omissions. The extracted predictions are not manually modified before scoring against the clinical reference annotations. Both original and normalized reports are evaluated against the same reference using the same NLG metrics. Raw scores and normalized-minus-raw differences are reported together.

\subsection{Training Utility} Standard supervised fine-tuning is used to assess the utility of \textsc{PanDent} as a training resource. Comparisons between the fine-tuned models and their corresponding pretrained baselines examine improvements in report-level textual agreement and in the tooth-level and regional clinical findings extracted from the generated reports.

\subsection{Evaluations}
\label{sec:main-evaluation}
\TMItabref{tab:nlg_raw_results} and \TMItabref{tab:attr_detailed_results} present report-level textual agreement and attribute-level clinical consistency on the common $505$-case benchmark.

\paragraph{Closed- and Open-Source Models}
Closed-source models obtain the highest zero-shot mean attribute accuracies. GPT-6 Astra leads with $0.457$, followed by GPT-6.1 Sol ($0.450$) and GPT-6 Sol ($0.445$). Gemini 4 Argon reaches $0.431$. Open-source models generally perform less well: Qwen3-VL-235B-A22B achieves $0.342$, compared with $0.269$ and $0.286$ for its 4B and 8B variants. Qwen3.5 also improves from $0.276$ at 4B to $0.295$ at 9B, although differences across model families indicate that size alone does not determine dental reporting performance. All zero-shot models remain below the fine-tuned baselines.

\paragraph{Medical-Domain Models}
Most medical-domain models achieve approximately $0.09$--$0.12$ mean attribute accuracy. OralGPT is an exception at $0.401$, outperforming the other medical-domain baselines but remaining below GPT-6 Astra ($0.457$). These results suggest that general medical specialization alone does not ensure accurate interpretation of panoramic dental images.

\paragraph{Textual Agreement versus Clinical Accuracy}
Normalization substantially increases textual agreement for several models, including Llama-3.2-90B-Vision, MAIRA-2, MVL-RRG-1.0, and OralGPT, highlighting the sensitivity of lexical metrics to reporting format. Gains are not uniformly positive, supporting the joint presentation of raw scores and normalization deltas. Normalized report agreement reaches $0.487$ for GPT-6 Astra. OralGPT attains $0.458$, exceeding Gemini 4 Argon ($0.432$), but has lower mean attribute accuracy ($0.401$ versus $0.431$). This ranking difference illustrates that stronger textual agreement does not necessarily imply more accurate tooth-level and regional findings in the same generated reports.

\paragraph{Supervised Fine-Tuning}
Fine-tuned Qwen3-VL-4B attains $0.760$ report agreement and $0.621$ mean attribute accuracy; Qwen3.5-4B reaches $0.775$ and $0.638$. Both improve over their corresponding zero-shot baselines without increasing model capacity. Relative to OralGPT, fine-tuned Qwen3.5-4B improves normalized textual agreement by $69.2\%$ and mean attribute accuracy by $59.2\%$, and also exceeds GPT-6 Astra ($0.487$ and $0.457$, respectively). These results demonstrate that \textsc{PanDent} provides effective supervision for learning the correspondence between panoramic image evidence, tooth-level findings, and clinical language, narrowing the performance gap between general-purpose MLLMs and expert dental interpretation.

\subsection{Case Study and Discussion}
\label{sec:case-study}
\TMItabref{tab:case_study} presents a qualitative comparison of generated reports against the reference findings for an in-house case, using selected output excerpts and reference-based analyses. GPT-5.4 recognizes generalized bone loss and root-canal treatment but under-specifies missing-tooth locations and omits non-metallic restorations.
Zero-shot Qwen3-VL-4B instead describes largely normal dentition, contradicting several positive findings. OralGPT produces a structured dental-style report but introduces extensive false-positive caries and fails to recover reference attributes. The fine-tuned model recovers non-metallic restorations, root-canal treatments, and generalized crestal bone loss, and identifies four of six missing teeth; errors remain at \#26 and \#27.

The case illustrates three observations consistent with the quantitative results. First, fluent or professionally structured reports may disagree with expert findings, so textual plausibility cannot substitute for clinical evaluation. Second, improvement after fine-tuning demonstrates the value of supervision that explicitly links panoramic evidence to dental attributes and their textual realization. Third, tooth-level grounding remains a source of residual error: a model may recognize a finding while missing individual FDI locations. More explicit tooth-aware visual grounding is therefore an important direction for improving clinical consistency.

\section{Conclusion}
This paper introduces \textsc{PanDent}, a large-scale panoramic dental benchmark emphasizing tooth-level reasoning and structure--language consistency. Clinically useful reporting requires accurate correspondence between radiographic evidence, tooth locations, and clinical findings, not only plausible descriptions. The benchmark combines expert-validated structured annotations, clinician-guided structured-to-text construction, multi-stage verification, and complementary textual and diagnostic evaluation. Experiments show that current models can generate fluent reports while making substantial tooth-level errors. Fine-tuning on \textsc{PanDent} improves both report alignment and diagnostic accuracy, demonstrating the value of fine-grained dental supervision and supporting clinically grounded multimodal evaluation and training.

\textbf{Limitations.}
Rare findings have limited positive support, increasing uncertainty in evaluation and learning. Annotation coverage and reproducibility also vary, particularly for subjective radiographic criteria. Implant-related prostheses are not explicitly modeled because of heterogeneous designs and limited subtype representation, although they are generally conspicuous on OPGs. Future work will broaden multi-source and multi-rater annotation, expand coverage of rare and implant-related findings, and improve tooth-aware grounding.

\section*{Acknowledgment}
GPT-Image-2 (OpenAI) generated the author-defined taxonomy illustration (\TMIfigref{fig:dataset_illustration}), reviewed by dental clinicians for clinical plausibility. No generative AI created or altered primary radiographs or research data images. The authors verified accuracy and completeness and take full responsibility.

\section*{References}
\IEEEtriggercmd{\balance\interlinepenalty=0\clubpenalty=0\widowpenalty=0}
\IEEEtriggeratref{9}
{
\small
\bibliographystyle{IEEEtran}
\bibliography{refs}
}
\end{document}